%% file: arxiv.tex
\documentclass[pageno]{asplos19-latex-template/jpaper_arxiv}

\usepackage{mathtools, amsfonts, amssymb}
\usepackage[english]{babel}
\PassOptionsToPackage{hyphens}{url}\usepackage{url}
\usepackage{tabularx}
\usepackage{booktabs}
\usepackage[ruled,linesnumbered]{algorithm2e}
\usepackage{xspace}
\usepackage{authblk}
\newcommand{\bF}{\mathbb{F}}
\graphicspath{{figs/}}

\input{macros}

\begin{document}

\title{CHET: Compiler and Runtime for Homomorphic Evaluation of Tensor Programs}

\author[*]{Roshan Dathathri}
\author[$\dagger$]{Olli Saarikivi}
\author[$\dagger$]{Hao Chen}
\author[$\dagger$]{Kim Laine}
\author[$\dagger$]{Kristin Lauter}
\author[$\dagger$]{Saeed Maleki}
\author[$\dagger$]{Madanlal Musuvathi}
\author[$\dagger$]{Todd Mytkowicz}
\affil[*]{Department of Computer Science, University of Texas at Austin, USA}
\affil[ ]{\textit {roshan@cs.utexas.edu}}
\affil[$\dagger$]{Microsoft Research, USA}
\affil[ ]{\textit {\{olsaarik,haoche,kilai,klauter,saemal,madanm,toddm\}@microsoft.com}}
\date{}
\maketitle
\thispagestyle{empty}

\input{sections/abstract}
\input{sections/introduction}
\input{sections/background}
\input{sections/stack}

\input{sections/hisa}
\input{sections/runtime}
\input{sections/compiler}
%\input{sections/optimizations}
\input{sections/evaluation}
\input{sections/related}
\input{sections/conclusion}

\bibliographystyle{plain}
\bibliography{references}

\end{document}

%% file: macros.tex
\newcommand{\Chet}{CHET}
\newcommand{\Hec}{\Chet{}~}
\newcommand{\Hecend}{\Chet}

\newcommand{\dtype}[1]{\operatorname{dtype}(#1)}
\newcommand{\shape}[1]{\operatorname{shape}(#1)}

\newcommand{\type}[1]{\texttt{#1}}

\newcommand{\void}{\texttt{void}}

\newcommand{\listtype}[2]{#1^{#2}}
\newcommand{\listlit}[1]{[#1]}

\newcommand{\pt}{\type{pt}} % slots
\newcommand{\ct}{\type{ct}} % slots

\newcommand{\hi}[1]{\textrm{#1}}
\newcommand{\hiencode}{\hi{encode}}
\newcommand{\hidecode}{\hi{decode}}
\newcommand{\hiencrypt}{\hi{encrypt}}
\newcommand{\hidecrypt}{\hi{decrypt}}
\newcommand{\hicopy}{\hi{copy}}
\newcommand{\hifree}{\hi{free}}
\newcommand{\hiadd}{\hi{add}}
\newcommand{\hiaddassign}{\hi{addAssign}}
\newcommand{\hiaddv}{\hi{addPlain}}
\newcommand{\hiaddvassign}{\hi{addPlainAssign}}
\newcommand{\hiadds}{\hi{addScalar}}
\newcommand{\hiaddsassign}{\hi{addScalarAssign}}
\newcommand{\hisub}{\hi{sub}}
\newcommand{\hisubassign}{\hi{subAssign}}
\newcommand{\hisubv}{\hi{subPlain}}
\newcommand{\hisubvassign}{\hi{subPlainAssign}}
\newcommand{\hisubs}{\hi{subScalar}}
\newcommand{\hisubsassign}{\hi{subScalarAssign}}
\newcommand{\himul}{\hi{mul}}
\newcommand{\himulassign}{\hi{mulAssign}}
\newcommand{\himulv}{\hi{mulPlain}}
\newcommand{\himulvassign}{\hi{mulPlainAssign}}
\newcommand{\himuls}{\hi{mulScalar}}
\newcommand{\himulsassign}{\hi{mulScalarAssign}}
\newcommand{\hirotl}{\hi{rotLeft}}
\newcommand{\hirotlassign}{\hi{rotLeftAssign}}
\newcommand{\hirotr}{\hi{rotRight}}
\newcommand{\hirotrassign}{\hi{rotRightAssign}}

\newcommand{\hidivs}{\hi{divScalar}}
\newcommand{\hidivsassign}{\hi{divScalarAssign}}
\newcommand{\himaxsdiv}{\hi{maxScalarDiv}}
\newcommand{\himulnorelin}{\hi{mulNoRelin}}
\newcommand{\himulnorelinassign}{\hi{mulNoRelinAssign}}
\newcommand{\hirelin}{\hi{relinearize}}
\newcommand{\hibootstrap}{\hi{bootstrap}}

%% file: sections/abstract.tex
\begin{abstract}
Fully Homomorphic Encryption (FHE) refers to a set of encryption schemes
that allow computations to be applied directly on encrypted data without
requiring a secret key. This enables novel application scenarios where a
client can safely offload storage and computation to a third-party
cloud provider  without having to trust the software and the
hardware vendors with the decryption keys. Recent advances in both FHE 
schemes and implementations have moved such applications from theoretical 
possibilities into the realm of practicalities.

This paper proposes a compact and well-reasoned interface called the
Homomorphic Instruction Set Architecture (HISA) for developing FHE
applications. Just as the hardware ISA interface enabled hardware
advances to proceed independent of software advances in the compiler
and language runtimes, HISA decouples compiler optimizations and
runtimes for supporting FHE applications from advancements in the
underlying FHE schemes.

This paper demonstrates the capabilities of HISA by building an
end-to-end software stack for evaluating neural network models on
encrypted data. Our stack includes an end-to-end compiler, runtime,
and a set of optimizations. Our approach shows generated code, on 
a set of popular neural network architectures, is  
faster than hand-optimized implementations.  
%% Homomorphic encryption (HE) refers to a family of encryption schemes that allow computations to be performed on encrypted data.
%% %Something about ML being our target?
%% Ensuring correctness of computations on HE requires selecting  precisions and encryption parameters that are large enough. On the other hand, for maximum efficiency these parameters should be kept as small as possible.
%% %, which the evaluation of machine learning models must be lowered to.
%% %However, commonly used operations such as ReLUs are not polynomial and must be replaced with approximations.
%% Furthermore, state-of-the-art \emph{leveled} HE schemes support SIMD multiplication, addition and rotations on large vectors of integers, which makes appropriate selection of data layouts and computational kernels critical for performance. These and other concerns make programming against HE libraries challenging.

%% We view HE libraries akin to a new piece of hardware, where the API corresponds to an ISA. This provides a crucial abstraction, that decouples optimization work on compilers and HE library implementations. We present an end-to-end compiler and a set of optimizations for evaluating neural network models on encrypted data.
%% %transforms to HE compatible with retraining
%% %optimizations
%% %parallel runtime
% %We evaluate our compiler on a set of off-the-shelf neural network architectures and show how our optimizations reach and exceed the performance of hand-written implementations.
\end{abstract}

%% file: sections/introduction.tex
\section{Introduction}
Many applications benefit from storing and computing on data in the
cloud. Cloud providers centralize and manage compute and storage
resources, which lets developers focus on the logic of their
application and not on how to deploy and manage the infrastructure
that surrounds it. A large class of applications, however, have
strict data secrecy and privacy requirements ranging from government 
regulations to gaining competitive advantage from consumer
protection. In these cases, data secrecy and privacy issues end up 
limiting otherwise convenient cloud adoption.

Fully Homomorphic Encryption~(FHE) allows computations to be applied
directly on encrypted data and thus enables a wide variety of cloud
applications that may not be available today. A cloud based FHE 
application \emph{encrypts} its \emph{plaintext} input $a$ and stores 
it in encrypted form in cloud storage as~$\alpha$. At a later time,
the application requests the cloud to perform an FHE operation $f$ on 
the encrypted data $\alpha$ and to store the (still encrypted) result 
$\beta = f(\alpha)$. The application can download $\beta$ from cloud 
storage at any point and \emph{decrypt} the result to obtain 
a plaintext result~$b = f(a)$.

FHE provides a simple trust model: the owner of the data does not
need to trust the cloud software provider or the hardware vendor.
Homomorphic encryption schemes rely on well-studied mathematical
hardness assumptions, i.e. if there is a polynomial time algorithm
for breaking the encryption then there is also a polynomial time
algorithm for solving these hard mathematical problems.
Other cryptographic solutions such as Secure Multi-Party Computation 
(MPC)~\cite{secureml,RouhaniRK17} require complicated trust 
models and typically have larger communication costs. Non-cryptographic 
technologies such as \emph{secure enclaves}~\cite{enclaves} (such as 
Intel SGX~\cite{sgx}) requires one to trust a hardware vendor.

%%% Explain the problems that motivate the need for a compiler
Of course, there is no free lunch: FHE is usually considered
impractical due to its performance overhead. However, recent advances
in FHE schemes and implementations dramatically improve performance,
which makes many applications practical today. For example, initial
FHE schemes~\cite{gentry2009fully} only operated on bits, requiring logic 
circuits to perform arithmetic. Modern FHE schemes~\cite{bgv,FV,heaan} 
organically support integer and fixed-precision arithmetic dramatically 
reducing the size of the circuits required. Similarly, optimized
implementations of FHE schemes such as~\cite{seal} have further reduced the
cost of performing individual FHE operations.

Despite these improvements, FHE operations are still several orders of
magnitude slower than their plaintext counterparts. Nevertheless,
certain privacy-sensitive applications are willing to pay the
cost. For instance, a bank based in Europe is unlikely to use
secure enclaves for sensitive computation as that requires trusting
Intel, an US company. For such applications, it is important to build
FHE applications that are as efficient as feasible, despite plaintext
overhead. 

Today, building an efficient FHE
application still requires manual and error-prone work. FHE encryption
schemes usually require a careful setting of encryption parameters
that trade off performance and security. Moreover, current FHE schemes
introduce noise when operating on encrypting data. Managing the growth
of this noise while satisfying the precision requirements of the
application currently requires laborious effort from the developer.
Also, popular FHE schemes support vectorized operations (called {\em
  batching} in the FHE literature) with very large vector sizes (of
thousands). Application efficiency crucially relies on maximizing the
parallelism available in the application to use the vector
capabilities. Lastly, FHE schemes limit the operations allowed in an
application (i.e., no branches or control flow) and thus require a
developer to creatively structure their program to work around such
limitations.

%%% Introduce HISA
In many respects, programming FHE applications directly on FHE schemes
today is akin to low-level assembly programming on modern
architectures. Our central hypothesis behind this paper is future FHE
applications will benefit from a compiler and runtime that targets a
compact and well-reasoned interface, which we call the Homomorphic
Instruction Set Architecture~(HISA). Just as the hardware ISA interface enabled
hardware advances to proceed independent of software advances in the
compiler and language runtimes, we posit that HISA isolates future
advances in FHE schemes from the compiler optimizations and runtime
improvements for supporting FHE applications. 

To demonstrate and evaluate the utility behind such an interface, this
paper focuses on the task of fully-homomorphic deep-neural
network~(DNN) inference. This application is particularly suited as
the first domain of study. First, the DNN inference computation does
not involve discrete control flow which is a perfect match for
FHE. While DNN inference involve non-polynomial activation functions,
they can be effectively replaced with polynomail
approximations~\cite{cryptonets,Chabanne2017}. The ability of DNNs to
tolerate errors provides us the flexibility of quantizing and reducing
the precision of the computation for performance. 

%% Introduce the compiler, the runtime, and the application
%On top of HISA, this paper presents a compiler and runtime called \Hec for
%developing FHE applications.  We evaluate HISA's efficacy by
%demonstrating it is a good substrate to develop a host of machine
%learning (ML) applications.

%% Explain what the compiler helps with
This paper presents, to the best of our knowledge, the first compiler
and runtime called \Hec for developing FHE DNN inference. \Hec
guarantees sound semantics by selecting the right
encryption parameters that maximize performance while guaranteeing
security and application-intended precision of the computed
results. \Hec builds on the recent encyrption scheme~\cite{heaan} that
provides approximate arithmetic to emulate fixed-point arithmetic
required for DNN inference. Finally, \Hec explores a large
space of optimization choices, such as the multiple ways to pack
data into ciphertexts as well as multiple ways to implement computational
kernels for each layout, that are currently done manually by experts today. 

%A compiler can help explore this space of optimizations
%to find good evaluation strategies with minimal user effort and expertise.

%Having a compiler helps
%guarantee proper semantics as well as security, as the encryption
%parameters for FHE schemes must be high enough to ensure both that
%computations succeed and security is maintained, while at the same
%time as low as possible for good performance. Furthermore, a recent
%promising encryption scheme provides approximate computation in
%exchange for radically better encryption parameters~\cite{heaan}. This
%opens up the possibility of the compiler helping to choose the
%smallest precision parameters possible, further improving performance.

%% Explain the runtime
In addition to a compiler, \Chet{} includes a runtime, akin to the linear algebra
libraries used in unencrypted evaluation of neural networks. We have developed a
set of layouts and a unified metadata representation for them. For these new
layouts we have developed a set of computational kernels that implement the
common operations found in convolutional neural networks (CNNs). All of these
kernels were designed to use the SIMD-style capabilities of modern FHE schemes.

%% Experiments

We evaluate \Hec with a set of real-world CNN models and show how different
optimization choices available in our compiler can significantly improve
latencies. We further demonstrate how the optimized kernel implementations in
\Hec allow neural networks of unprecedented depth to be homomorphically
evaluated.

%% summarize

In summary our main contributions are as follows:
\begin{itemize}
\item A homomorphic instruction set architecture (HISA) that cleanly abstracts
and exposes features of FHE libraries.
\item A set of tiled layouts and a metadata format for packing tensors into ciphertexts.
\item Novel kernel implementations of homomorphic tensor operations.
\item A compiler and runtime for homomorphic evaluation of tensor programs, known as \Chet{}.
\item An evaluation of \Chet{} on a set of real-world CNN models, including the
homomorphic evaluation of the deepest CNN to date.
\end{itemize}

%The rest of this paper is organized as follows: Section~\ref{sec:background} introduces important background and definitions, Section~\ref{sec:hisa} describes the HISA, Section~\ref{sec:hec-stack} presents the layouts, kernels, compiler and runtime, Section~\ref{sec:eval}

%% file: sections/background.tex
\section{Background}
\label{sec:background}
This Section provides a concise background of homomorphic encryption with of its
limitations and capabilities, followed by a brief introduction into tensor
programs.

\subsection{Homomorphic Encryption}

We will begin by describing the properties of \emph{leveled} fully homomorphic
encryption (FHE) that are essential for a programmer. We first introduce general
concepts followed by a description of the HEAAN scheme~\cite{heaan}, which is
used in our evaluations.
%, followed by how the\emph{bootstrapping} operation permits true \emph{Fully
%Homomorphic Encryption} and what the concerns are for using it.

Leveled FHE can evaluate an arithmetic circuit of a pre-determined
multiplicative depth~$L$. The reason for such a restriction is that typically in
homomorphic encryption each ciphertext carries a inherent ``noise'' which
increases when computations are performed. Once the noise reaches a certain
bound the ciphertext becomes corrupted and the message cannot be recovered
anymore, even with the correct decryption key. In Leveled FHE, the parameters of
the encryption scheme are chosen so that the noise remains below this threshold
for circuits of depth up to the chosen value~$L$. Unfortunately, due to the
increase in the parameters of the scheme, the complexity of each homomorphic
operation also grows with~$L$.

\subsubsection*{Encryption Keys:}
FHE schemes have two sets of keys: {\em public keys} and {\em private keys}.
Private keys are used for decryption and public keys are used for encryption and
in homomorphic operations. As long as a user does not reveal their private key,
FHE schemes leak no information about the user's data.\footnote{Encryption
schemes in general base their security on the hardness of some underlying
problem, e.g., learning with errors~\cite{lwe}.} Public keys, on the other hand,
are meant to be shared for secure computation.

\subsubsection*{Supported Operations:}
\label{sec:func}
Existing FHE schemes on integers support only two basic algebraic operations:
{\em addition} and {\em multiplication}. The operands can be any
combination of ciphertexts and plaintexts. Techniques for fixed point operands
are discussed later in this Section. However, all other operations need to be
approximated in terms of additions and multiplications. For example, functions
such as $\exp$, $\log$, and $\tanh$ can be approximated by polynomials using
Taylor expansion. On the other hand, branching is not possible in FHE and the
only possible solution is executing all cases and masking the result.

\subsubsection*{Noise and Multiplicative Depth:}
We denote the amount of {\bf noise} associated with ciphertext $\alpha$ by
$noise(\alpha)$. The amount of noise for a ciphertext depends on its operands
and the operator and excessive noise completely destroys the encrypted message.
Any freshly encrypted value has a small amount of noise. For any given
ciphertexts, $\alpha$ and $\beta$ and plaintext $c$, the noise contribution
is roughly as follows:
\begin{itemize}
\item $noise(\alpha \times c) \approx noise(\alpha + c) \approx noise(\alpha)$
\item $noise(\alpha + \beta) \approx \max\{noise(\alpha), noise(\beta)\}$
\item $noise(\alpha \times \beta) \approx \max\{noise(\alpha), noise(\beta)\} \cdot r$
\end{itemize}
where $r$ is a constant number that depends on parameter selection of the
encryption. The multiplicative noise increase between two ciphertexts greatly
affects the design of a FHE computation. {\em Multiplicative depth} of a
ciphertext $\alpha$ is the maximum number of ciphertext multiplications that
contributed to the computation of $\alpha$ and they are all a part of a chain.
For example, for ciphertext $\alpha = \beta * (\gamma * \gamma + a + \lambda)$
where $\beta$, $\gamma$ and $\lambda$ are ciphertexts and $a$ is a plaintext,
the multiplicative depth is $2$. For a multiplicative depth of $L$, the noise is
$O(r^L)$. The maximum noise tolerance can be adjusted by the encryption
parameters but that negatively impacts computational complexity of operations.
Therefore, it is crucial to keep the multiplicative depth minimal to avoid the
costs.
 
\subsubsection*{Floating Point:}
\label{sec:floats}
Due to the nature of the homomorphic encryption schemes, it is not easy to
support floating point operations. Instead, we can use fixed point arithmetic by
combining integers with a scaling factor.  To avoid accumulating scaling factor,
fixed point arithmetic requires scaling the result down after each
multiplication.  However, the BFV/BGV schemes do not support division, and hence
applications using these schemes will see an accumulation of scaling factors.
This requires increasing space to store the intermediate results, and the
required encryption parameters becomes exponential in the multiplicative depth
of the circuit, which makes deep circuit evaluations infeasible.
Section~\ref{sec:heaan} gives background on an \emph{approximate} encryption
scheme that avoids this problem.

\subsubsection*{FHE Vectorization:}
FHE multiplication and addition are extremely costly and usually not practical
if one pair of operands are considered at a time. Fortunately, in the BFV~\cite{FV}, BGV~\cite{bgv}
and HEAAN schemes, multiple ciphertexts can be packed into a single ciphertext
to benefit from Single Instruction Multiple Data (SIMD)
parallelism~\cite{flynntaxonomy}. This technique allows packing multiple
plaintext values into a single ciphertext. Given two $\alpha$ and $\beta$, that
encode the plaintext vectors $(a_1,a_2,\dots,a_n)$ and $(b_1,b_2,\dots,b_n)$,
respectively, $\alpha\times\beta$ or $\alpha+\beta$ correspond to elementwise
multiplication and addition of the underlying individual $a_i$ and $b_i$.
Similar to shuffling instructions in the SIMD units of modern processors, the
elements of a vector can be rotated in a packed ciphertext. However, random
access is not directly supported. Instead random access can be implemented by
multiplying with a plaintext mask, followed by a rotation. This unfortunately
introduces additional multiplicative depth and should thus be avoided if
possible.

\subsubsection*{Bootstrapping:}
True Fully Homomorphic Encryption starts with a Leveled FHE scheme and
introduces an additional noise management technique traditionally referred to as
bootstrapping. Bootstrapping reduces noise, and when performed frequently enough
it allows in principle an unlimited number of operations to be performed. A
recent work by Cheon et al.~\cite{heaanboot} describes a bootstrapping algorithm
for the HEAAN~\cite{heaan} scheme, and best results takes about 1 second to
bootstrap a single number. Therefore, we leave using bootstrapping for future
work once it is more practical.

%It is a map {\em bootstrap} in the space of all possible ciphertexts satisfying two properties. First,if $c$ decrypts to 
%a message $m$, then $c' = \textsf{bootstrap}(c)$ also decrypts to $m$. Moreover, if the underlying leveled FHE scheme 
%supports evaluation of depth-$L$ circuits on fresh ciphertexts, and the \textsf{bootstrap} procedure can be expressed a
%s a depth $D$ circuit with $D < L$, then the output ciphertext $c'$ has $L-D$ levels left. Hence, we could 
%periodically perform bootstrapping to enable arbitrary depth computation on encrypted data. 
%Bootstrapping is generally considered an expensive operation. A recent work by Cheon et al. \cite{heaanboot} describes a 
%bootstrapping algorithm for the HEAAN scheme, and  best results takes about 1 second to bootstrap a single number. 
%Note that the HEAAN bootstrapping only outputs approximate results, i.e., the output ciphertext $c'$ decrypts to 
%a message $m' = m+e$ for some small error $e$.  This is due to the fact that
%this scheme only supports approximate computation. 

\subsection{HEAAN}
\label{sec:heaan}

There are a few state-of-the-art leveled FHE schemes, for example the
Brakerski-Gentry-Vaikuthanathan scheme \cite{bgv} and the Fan-Vercauteren
variant of the Brakerski scheme \cite{FV}, which we will refer to as BGV and
BFV. These two schemes are similar in many aspects. In particular, both scheme
supports vector plaintext type $\bF^k$ where $\bF$ is some finite field, and
vectorized additions/multiplications. More recently, Cheon et al. proposed the
HEAAN scheme~\cite{heaan}, which supports better approximate computation over
encrypted data.

The HEAAN scheme  uses a novel idea to mix the encryption noise with the
message. Then, one can use the modulus switching techniques for BGV scheme to
achieve the functionality of ``scaling down'', i.e., given an encryption of $m$
and a public integer $K$, output an encryption of $m/K + \epsilon$ for some
$\epsilon > 0$ of fixed size. This gives the ability to control the scaling
factor and hence the precision throughout the homomorphic circuit evaluation,
and allows encryption parameters that are linear in multiplicative depth.
However, the HEAAN scheme only supports approximate computations: first, even a
freshly encrypted ciphertext decrypts to a perturbed message; also, homomorphic
operations insert noise into the result. Therefore, one needs to carefully
control these noises in order to maintain a desired precision of the computation
output. 

In the HEAAN scheme, two operations require additional public key materials.
After each multiplication between ciphertexts, the output ciphertext is actually
larger than the two input ciphertexts, and it's common to run an operation
called \emph{re-linearization} to reduce the size of output. Both
re-linearization and rotation require certain evaluation keys. These keys can be
generated from the secret key, and each rotation key can perform rotation by a
fixed amount. In the public implementation of HEAAN, by default only power-of-2
rotation keys are generated, and general rotations are written as a composition
of power-of-2 rotations. This presents an interesting problem of choosing the
best rotation keys for a given program, which balances evaluation time with
storage/communication costs of additional rotation keys.

To instantiate the HEAAN scheme, we need to select two integer
parameters $N$ (the degree of polynomial modulus, a power of 2) and $Q$ (the
ciphertext modulus). The security of the scheme is based on the ring
learning-with-errors problem, and for a fixed $Q$,  the security level increases
with $N$. One additional complication is: the noise generated by each
homomorphic operation increases with $N$, which could increase required
precision and thus $Q$, possibly requiring a larger $N$ than the initial guess.

The number of elements that can be packed into a single ciphertext scales
linearly with the parameter $N$. In particular, HEAAN supports encrypting $N/2$
complex numbers in a single ciphertext. We note that the vector encryption in
HEAAN comes with an additional encoding error, i.e., even a fresh encoding of a
vector will lose some precision of the input due to a rounding process. In
HEAAN, the encoding error is $O(\sqrt{N})$, hence we need to scale up the input
numbers to make sure they are sufficiently larger compared to this error. 

%\subsection{HEAAN Operation Costs}
%
%\begin{table}
%\begin{tabular}{ | c | c | }
%  Operation  & Cost \\
%  \hline   \hline 
%  \verb|add|, \verb|sub| & $O(n)$ add/sub \\ 
%   \verb|mulScalar|, \verb|rescale| & $O(n)$ mult \\
%  \verb|mul|, \verb|mulPlain| & $O(n\log(n)) $ mult \\
%  \verb|relinearize|, \verb|rotLeft|, \verb|rotRight| & $O(n \log (n))$ mult.  
%\end{tabular}
%\caption{HEAAN asymptotic costs. Unit: operations in integers mod $Q$}
%\end{table}

%\subsubsection*{Side note}
%
%We note that the cost structure of homomorphic operations could vary a lot based on small variations of the scheme. For example, by using a full-RNS variant of the HEAAN scheme and storing the ciphertexts and plaintexts in the number-theoretic-transform (NTT) domain, we can decrease the complexity of \verb|multByPoly| from $O(n\log(n))$ to $O(n)$. The goal for our homomorphic compiler is to be able to recognize this change and make best choices accordingly.

\subsection{Tensor Programs}

A tensor is a multidimensional array with regular dimensions. A tensor $t$ has a
data type $\dtype{t}$, which defines the representation of each element, and a
shape $\shape{t}$, which is a list of the tensor's dimensions. For example, a
single 32 by 32 image with 3 channels of color values between 0 and 255 could be represented by a tensor $I$
with $\dtype{I}=\type{int8}$ and $\shape{I}=\listlit{3,32,32}$.

In machine learning, neural networks are commonly expressed as programs
operating on tensors. Some common tensor operations in neural
networks include:
\begin{description}
\item[Convolution] calculates a cross correlation between an input and a filter tensor.
\item[Matrix multiplication] is used to represent neurons that are connected to
all inputs, which are found in dense layers typically at the end of a
convolutional neural network.
\item[Pooling] combines adjacent elements using a reduction operation such as maximum or average.
\item[Elementwise operations] such as ReLUs and batch normalization.
\item[Reshaping] reiterprets the shape of a tensor, for example, to flatten
preceding a matrix multiplication.
\end{description}

Tensor programs are used as models for large datasets in regression and
classification tasks. To define a space of models some constant tensors in the
program are designated as trainable weights. Since operations are
differentiable (almost everywhere), backpropagation can be used to update
weights. This is used as a step in an optimization algorithm, e.g., stochastic
gradient descent.

Tensor programs have attractive properties for execution on homomorphic
encryption: typically no branching, very regular data access patterns making
vectorization possible. We consider the tensor program as a circuit of tensor operations and this circuit is a Directed Acyclic Graph (DAG).

%% file: sections/stack.tex
% !TEX root = ../main.tex
\section{Software Stack for Homomorphic Evaluation of Tensor Programs}
\label{sec:hec-stack}

\begin{figure}
\centering
\includegraphics[width=0.35\textwidth]{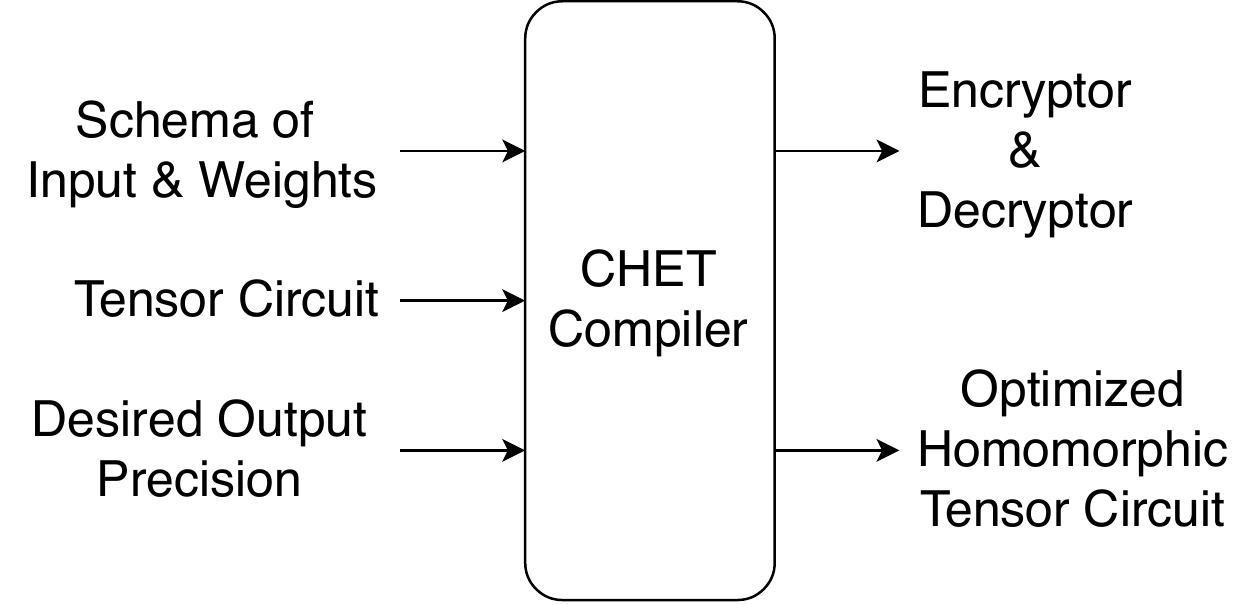}
\caption{Overview of the \Hec system at compile-time.}
\label{fig:overview-compiler}
\end{figure}

\begin{figure*}
\centering
\includegraphics[width=0.99\textwidth]{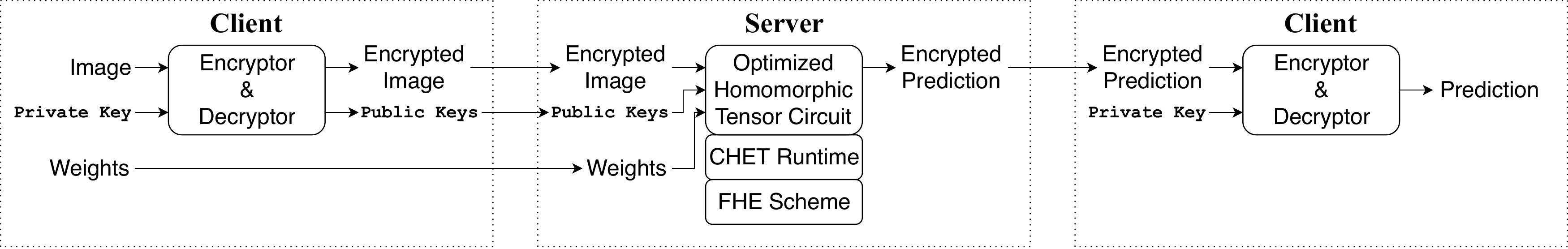}
\caption{Overview of the \Hec system at runtime.}
\label{fig:overview-runtime}
\end{figure*}

There are many open problems in building a full software stack for FHE applications and the systems, OS, and PL community can help bridge the gap between an application programmer and the underlying crypto library that implements secure primitives.  This section presents an overview of an end-to-end software stack for evaluating \emph{tensor programs} on homomorphically encrypted data. 
The target architecture for this stack is a Fully Homomorphic Encryption (FHE) scheme and our tensor compiler, \Hecend, interacts with this scheme using the Homomorphic Instruction Set Architecture (HISA) defined in Section~\ref{sec:hisa}. 
At the top of stack rests the user-provided tensor program or circuit. 
\Hec consists of a compiler and runtime to bridge the gap in-between them. 
We describe the runtime and the compiler in detail in Sections~\ref{sec:hec-runtime} and~\ref{sec:hec-compiler}, respectively.
In this Section, we present an overview of the end-to-end software stack.

% \subsection{Overview}
% \label{sec:hec-overview}

\input{sections/overview}

%% file: sections/overview.tex
% \paragraph{CHET:} 
Figure~\ref{fig:overview-compiler} shows the overview of the \Hec compiler. 
In addition to the tensor circuit, \Hec requires the schema of the input and weights to the circuit. The schema specifies the dimensions of the tensors as well the floating-point precision required of the values in those tensors. \Hec also requires the desired floating-point precision of the output of the circuit. Using these constraints, \Hec generates an equivalent, optimized homomorphic tensor circuit as well as an encryptor and decryptor. Both of these executables encode the choices made by the compiler to make the homomorphic computation efficient. 

To evaluate the tensor circuit on an image, the client first generates a private key and encrypts the image using the encryptor (which can also generate private keys) provided by the compiler, as shown in Figure~\ref{fig:overview-runtime}. The encrypted image is then sent to the server along with unencrypted weights and public keys required for evaluating homomorphic operations (i.e., multiplication and rotation). The server executes the optimized homomorphic tensor circuit generated by the \Hec compiler. The homomorphic tensor operations in the circuit are executed using the \Hec runtime, which uses an underlying FHE scheme to execute homomorphic computations on encrypted data. The circuit produces an encrypted prediction, which it then sends to the client. The client decrypts the encrypted prediction with its private keys using the compiler generated decryptor.  In this way, the client runs tensor programs like neural networks on the server without the server being privy to the data, the output (prediction), or any intermediate state.

While Figure~\ref{fig:overview-runtime} presents the flow in \Hec for homomorphically evaluating a tensor program on a single image, \Hec supports evaluating the program on multiple images simultaneously, which is known as \textit{batching} in image inference. Batching increases the throughput of image inference. In contrast, \Hec's focus is decreasing image inference latency. In the rest of this paper, we consider a batch size of 1, although \Hec trivially supports larger batch sizes.

The FHE scheme exposes several policies or heuristics such as the encryption parameters. Similarly, the \Hec runtime also exposes several policies that could be specific to or independent of the FHE scheme. This is analogous to Intel MKL libraries having different implementations of the same operation, where the most performant implementation depends on the size of the input or the target architecture. In the case of homomorphic computation, these policies not only affect performance but can also affect security and accuracy. A key design principle of \Hec is the separation of concern between the policies of choosing the secure, accurate, and most efficient homomorphic operation from the mechanisms of executing those policies. Some of these policies are independent of the input or weights schema, so they are entrusted to the \Hec runtime. On the other hand, most policies may require either the input or weights schema or global analysis of the program. In such cases, the \Hec compiler chooses the appropriate policy and the \Hec runtime implements the mechanisms for that policy. We describe the policies explored and mechanisms exposed by the \Hec runtime (Section~\ref{sec:hec-runtime}) and then describe how the \Hec compiler (Section~\ref{sec:hec-compiler}) chooses the appropriate policy.

%% file: sections/hisa.tex
\section{Homomorphic Instruction Set Architecture}
\label{sec:hisa}

This section introduces a design for a Homomorphic Instruction Set Architecture
(HISA), that provides a useful interface to fully homomorphic encryption
libraries. We have designed the HISA with the following objectives in mind:
\begin{itemize}
\item Abstract details of HE schemes, such as use of evaluation keys and
management of moduli.
\item Provide a common interface to shared functionality in FHE libraries, while
exposing unique features.
\item Not to include the complexity of plaintext evaluation. Instead the HISA is
intended to be embedded into a host language.
\end{itemize}

The HISA is split into multiple profiles and all libraries are expected to
implement at least the Encryption profile. Each FHE library implementing the
HISA provides two types, $\pt$ for plaintexts and $\ct$ for ciphertexts, and
additional ones depending on the profiles implemented by the library.

\begin{figure*}
\footnotesize
\begin{tabularx}{\textwidth}{@{}Xlp{6.2cm}r@{}}
    \toprule
Instruction & Signature & Semantics & Profile \\
\midrule
$\hiencrypt(p)$
    & $\pt \rightarrow \ct$
    & Encrypt plaintext $p$ into a ciphertext.
    & Encryption
    \\
$\hidecrypt(c)$
    & $\ct \rightarrow \pt$
    & Decrypt ciphertext $c$ into a plaintext.
    & Encryption
    \\
$\hicopy(c)$
    & $\ct \rightarrow \ct$
    & Make a copy of ciphertext $c$.
    & Encryption
    \\
$\hifree(h)$
    & $\ct\cup\pt \rightarrow \void$
    & Free any resources associated with handle $h$.
    & Encryption
    \\
\midrule
$\hiencode(m)$
    & $\listtype{\mathbb{Z}}{s} \rightarrow \pt$
    & Encode vector of integers $m$ into a plaintext.
    & Integers
    \\
$\hidecode(p)$
    & $\pt \rightarrow \listtype{\mathbb{Z}}{s}$
    & Decode plaintext $p$ into a vector of integers.
    & Integers
    \\
$\hirotl(c,x),\hirotlassign(c,x)$
    & $\ct,\mathbb{Z} \rightarrow \ct$
    & Rotate ciphertext $c$ left $x$ slots.
    & Integers
    \\
$\hirotr(c,x),\hirotrassign(c,x)$
    & $\ct,\mathbb{Z} \rightarrow \ct$
    & Rotate ciphertext $c$ right $x$ slots.
    & Integers
    \\
\midrule
$\hiadd(c,c'),\hiaddassign(c,c')$
    & $\ct,\ct \rightarrow \ct$
    & \multirow{3}{6.2cm}{Add ciphertext, plaintext, or scalar to ciphertext $c$.}
    & Integers
    \\
$\hiaddv(c,p),\hiaddvassign(c,p)$
    & $\ct,\pt \rightarrow \ct$
    &
    & Integers
    \\
$\hiadds(c,x),\hiaddsassign(c,x)$
    & $\ct,\mathbb{Z} \rightarrow \ct$
    &
    & Integers
    \\
    \midrule
$\hisub(c,c'),\hisubassign(c,c')$
    & $\ct,\ct \rightarrow \ct$
    & \multirow{3}{6.2cm}{Subtract ciphertext, plaintext, or scalar from ciphertext $c$.}
    & Integers
    \\
$\hisubv(c,p),\hisubvassign(c,p)$
    & $\ct,\pt \rightarrow \ct$
    &
    & Integers
    \\
$\hisubs(c,x),\hisubsassign(c,x)$
    & $\ct,\mathbb{Z} \rightarrow \ct$
    &
    & Integers
    \\
    \midrule
$\himul(c,c'),\himulassign(c,c')$
    & $\ct,\ct \rightarrow \ct$
    & \multirow{3}{6.2cm}{Multiply ciphertext $c$ with ciphertext, plaintext, or scalar.}
    & Integers
    \\
$\himulv(c,p),\himulvassign(c,p)$
    & $\ct,\pt \rightarrow \ct$
    &
    & Integers
    \\
$\himuls(c,x),\himulsassign(c,x)$
    & $\ct,\mathbb{Z} \rightarrow \ct$
    &
    & Integers
    \\
    \midrule
$\hidivs(c,x),\hidivsassign(c,x)$
    & $\ct,\mathbb{Z} \rightarrow \ct$
    & Divide ciphertext $c$ with scalar $x$. Undefined if $x\not=\himaxsdiv(c,\mathit{ub})$ for some $ub\in\mathbb{Z}$.
    & Division
    \\
$\himaxsdiv(c,\mathit{ub})$
    & $\ct,\mathbb{Z} \rightarrow \mathbb{Z}$
    & Gives largest valid divisor $d$ s.t. $1\leq d\leq \mathit{ub}$.
    & Division
    \\
    \midrule
$\himulnorelin(c,c'),\himulnorelinassign(c,c')$
    & $\ct,\ct \rightarrow \ct$
    & Ciphertext-ciphertext multiplication without re-linearization.
    & Relin
    \\
$\hirelin(c)$
    & $\ct \rightarrow \void$
    & No-op, FHE library performs re-linearization.
    & Relin
    \\
\midrule
$\hibootstrap(c)$
    & $\ct \rightarrow \void$
    & No-op, FHE library performs bootstrapping.
    & Bootstrap
    \\
\bottomrule
\end{tabularx}
\caption{Instructions of the HISA\label{fig-hisa-inst}}
\vspace{-8pt}
\end{figure*}

Figure~\ref{fig-hisa-inst} presents instructions available in the HISA. The
Encryption profile provides core functionality for encryption, decryption, and
memory management. Usage of encryption and evaluation keys is left as a
responsibility of FHE libraries and is not exposed in the HISA.

The Integers profile provides operations for encoding and computing on integers.
The profile covers both FHE schemes with batched encodings and ones without by
having a configurable number of slots $s$, which is provided as an additional
parameter during initialization of the FHE library.

The Division profile exposes extra functionality provided by the HEAAN
family of encryption schemes. For libraries implementing the base HEAAN
scheme~\cite{heaan}, $\himaxsdiv(c,\mathit{ub})$ will return the largest
power-of-two $d$ such that $d\leq \mathit{ub}$ and $d\leq q$, where $q$ is the
modulus of $c$. For a variation of HEAAN based on residue number systems
(RNS)~\cite{rnsfv}, on the other hand, $\himaxsdiv(c,\mathit{ub})$ would
return the largest coprime modulus of $c$ less than $\mathit{ub}$, or 1 if there is
none. This way the Division profile cleanly exposes the division functionality
of both variants of the HEAAN encryption scheme.

The Relin profile provides the capability to separate multiplication from
re-linearization. While re-linearization is semantically a no-op, it is
useful to expose $\hirelin$ calls to the compiler, as their proper placement
is a highly non-trivial (in fact NP-complete) problem~\cite{HaoOptRelin}.

Finally, the Bootstrap profile exposes bootstrapping in FHE libraries that
support it. While bootstrapping is semantically a no-op, it must be exposed in
the HISA because selection of encryption parameters depends on when
bootstrapping is performed. Furthermore, the optimal placement of bootstrapping
operations~\cite{benhamouda2017optimization} depends on the program being evaluated: ``wide'' programs with a lot
of concurrent should bootstrap at lower depths than ``narrow'' programs that are
mostly sequential.

% The HISA does not currently cover bitwise FHE libraries such as
% TFHE~\cite{TFHE}. For these the HISA should be extended with a new profile.

The HISA may be implemented either with precise or \emph{approximate} semantics.
With approximate semantics all operations that return a $\pt$ or $\ct$ may
introduce an error term. It is expected that this error is from some random
distribution which may be bounded given a confidence level. However, the HISA
does not offer instructions for estimating the error, as we expect these
estimates to be required during selection of fixed point precisions and
encryption parameters. Instead we recommend that approximate FHE libraries
provide implementations of the HISA with no actual encryption and either:
% \begin{enumerate}
% \item 
a way to query safe estimates of errors accumulated in each $\pt$ and
$\ct$, or
% \item 
sampling of error from the same distribution as encrypted evaluation would
produce.
% \end{enumerate}
Both approaches may be used by a compiler for selecting fixed point precisions.
The sampling approach is more flexible for applications where it is hard to
quantify acceptable bounds for error, such as neural network classification,
where only maintaining the order of results matters.

Initialization of FHE libraries is concerns that falls outside of the HISA, and
is specific to the encryption and encoding scheme used. This step will include
at least generating or importing encryption and evaluation keys. FHE libraries
implementing leveled encryption schemes also require selecting encryption
parameters that are large enough to allow the computation to succeed. Such FHE
libraries should provide analysis passes as additional implementations of the
HISA to help with parameter selection. We have implemented such a pass for the
HEAAN library~\cite{heaan}, which we present in Section~\ref{sec:hec-compiler}.

\paragraph{HISA is not an IR:}
The HISA does not and nor is it meant to provide a unified interface to FHE
libraries. To draw a parallel to processors implementing the x86 instruction
set, if a program includes instructions from Advanced Vector Extensions (AVX)
then that program will not run on any processor bought before 2011. Conversely,
running a program that does not use AVX on a processor that does may be leaving
a lot of potential performance unused. Similarly, a program using the Division
HISA profile will only run on FHE libraries implementing a HEAAN style
encryption scheme~\cite{heaan} and again, conversely, a program using HEAAN for
fixed point arithmetic without ever calling $\hidivs$ would likely achieve much
better performance (through smaller encryption parameters).
The role of a unified interface should instead be played by an intermediate
representation (IR). For the HISA, this IR would include fixed point datatypes
and specifications of required precisions. This kind of an IR could then be
lowered to target FHE libraries that support different subsets of profiles. This
is similar to how for example the LLVM intermediate representation can be
lowered both to processors that support AVX and ones that do not.

%The rest of this paper assumes that the FHE library supports the Core, Integers
%and Division profiles, but our techniques would work largely without
%modifications with just the Core and Integers profiles.

%% file: sections/runtime.tex
% !TEX root = ../main.tex
\section{Runtime for Executing Homomorphic Tensor Operations}
\label{sec:hec-runtime}

Intel MKL libraries provide efficient implementations of BLAS operations. In the same way, we design the \Hec runtime to provide efficient implementations for homomorphic tensor primitives. While the interface for BLAS and tensor operations are well-defined, the interface for a tensor operation on encrypted data is not apparent because the encrypted data is an encryption of a vector, whereas the tensor operation is on higher-dimensional tensors. The types need to be reconciled to define a clean interface. In the rest of this section, we consider a 4-dimensional tensor (batch, channel, height, and width dimensions) to illustrate this, but the same concepts apply to other higher-dimensional tensors. We first describe our cipher tensor datatype. Each tensor operation on unencrypted tensors corresponds to an equivalent homomorphic tensor operation on cipher tensor(s) and plain tensor(s), if any.
We then briefly describe implementations of some homomorphic tensor operations using this clean interface and datatype.

\subsection{Cipher Tensor Datatype:}

A naive way to encode a 4-d tensor as a vector is to lay out all elements in the tensor contiguously in the vector and encrypt it. This approach may not be feasible for a few reasons: 
% \begin{itemize}
% \item 
(i) the vector size is determined by the encryption parameters (like N in HEAAN) and the 4-d tensor might not fit in the vector, or
% \item  
(ii) tensor operations like convolution and matrix multiplication on this input vector (or to produce such an output vector) become complicated and very inefficient because only point-wise operations or rotations are supported by the HISA.
% \end{itemize} 
Another option is to encrypt each inner dimension (width) element in a separate cipher. This creates a vector of vectors, where the inner vector is a cipher and the outer vector contiguously stores pointers to these ciphers. The number of homomorphic operations would increase significantly because a cipher operation may be required for each element, thereby not utilizing the vector width and making it very inefficient. 

The problem of encoding a 4-d tensor as a vector of vectors is similar to tiling or blocking the data but with constraints that differ from that of locality. For example, a 4-d tensor can be blocked as a 2-d tensor of 2-d tensors. This corresponds to the 4-d tensor being laid out as a vector of vectors, where each inner vector has the inner two dimensions (height and width) encrypted. Similar issues as stated earlier may arise if a particular data layout is fixed. Some tensor operations might be more efficient in some layouts than in others. More importantly, the most efficient layout may be dependent on the schema of the tensor or on future tensor operations, so determining that is best left to the compiler. Thus, we want a uniform cipher tensor datatype that is parametric to different data layouts, so that the \Hec compiler can choose the appropriate data layout.

Some tensor operations enforce more constraints on the cipher tensor datatype, primarily for performance reasons.
For example, to perform convolution with same padding, the input 4-d tensor is expected to be padded. Such a padding on unencrypted data is trivial, but adding such padding to an encrypted cipher on-the-fly involves several rotation and point-wise multiplication operations. Similarly, a reshape of the tensor is trivial on unencrypted data but very inefficient on encrypted data. All these constraints arise because using only point-wise operations or rotations on the vector may be highly inefficient to perform certain tensor operations. Therefore, the cipher tensor datatype needs to allow padding or logical re-shaping without changing the elements of the tensor. 

% To summarize, the cipher tensor datatype needs to meet the following constraints:
% % \begin{itemize}
%     % \item 
%     (i) tiling of the data layout of the encrypted tensor should be parametric,
%     % \item 
%     (ii) the data layout should support padding empty or invalid elements in-between the elements of the tensor, and 
%     % \item 
%     (iii) the data layout should be distinct from the logical shape of the tensor.
% % \end{itemize}

To satisfy these requirements, we define an cipher tensor datatype, that we term \textsf{CipherTensor}, as a vector of ciphertexts (vectors) with associated metadata that captures the way to interpret the vector of ciphertexts as the corresponding unencrypted tensor. 
The metadata includes:
% \begin{itemize}
    % \item 
    (i) physical dimensions of the (outer) vector and those of the (inner) ciphertext, 
    % \item 
    (ii) logical dimensions of the equivalent unencrypted tensor, and 
    % \item 
    (iii) physical strides for each dimension of the (inner) ciphertext.
% \end{itemize}

The metadata is stored as plain integers and can be modified easily. Nevertheless, \textit{the metadata does not leak any information about the data because it is agnostic to the values of the tensor and is solely reliant on the schema or dimensions of the tensor}. The metadata can be used to satisfy all the required constraints of the datatype:
\begin{itemize}
    \item The metadata are parameters that the compiler can choose to instantiate and use a specific data layout. For example, blocking or tiling the inner dimension (height and width) of a 4-d tensor corresponds to choosing 2 dimensions for the (outer) vector and 2 dimensions for the (inner) ciphertext, where the logical dimensions match the physical dimensions. We term this as \textsf{HW-tiling}. Another option is to block the channel dimension too, such that multiple, but not all, channels may be in a cipher. This corresponds to choosing 2 dimensions for the (outer) vector and 3 dimensions for the (inner) ciphertext with only 4 logical dimensions. We term this as 
    \textsf{CHW-tiling}.
    \item The physical strides of each dimension can be specified to include sufficient padding in-between elements of that dimension. For example, for an image of height (row) and width (column) of 28, a stride of 1 for the width dimension and a stride of 30 for the height dimension allows a padding of 2 (zero or invalid) elements between the rows.
    \item Reshaping the cipher tensor only involves updating the metadata to change the logical dimensions and does not perform any homomorphic operations. % In some arbitrary cases, it might require more.
\end{itemize}

\input{sections/conv2d}

\subsection{Homomorphic Tensor Operations:}

The interface for a homormorphic tensor operation is mostly similar to that of its unencrypted tensor operation counterpart, except that the types are replaced with cipher tensors or plain tensors, which we call CipherTensor and PlainTensor, respectively. The homomorphic tensor operation also exposes the output CipherTensor in the interface as a pass-by-reference parameter. This enables the compiler to specify the data layout of both the input and output CipherTensors. In addition, the interface exposes parameters to specify the scaling factors to use for the input CipherTensor(s) and PlainTensor(s), if any. When a compiler specifies all these parameters, it chooses an implementation specialized for this specification. There might be multiple implementations for the same specification that are tuned for specific FHE libraries (for those that have \hidivs~and for those that do not). In such cases, the compiler also chooses the implementation to use. Nevertheless, there are several algorithm choices for a given specification that have significant impact on the performance of the operation. These could be quite involved (similar to MKL implementations). We explore some of these briefly next.

\paragraph{HW-tiled Homomorphic Convolution 2-d with VALID padding}: 
Consider a HW-tiled CipherTensor that represents a 4-d tensor. Algorithm~\ref{alg:conv2d} shows pseudocode for a 2-d convolution with valid padding of this CipherTensor into another HW-tiled CipherTensor. Zeros (of the size of the ciphertext) encrypted into a ciphertext is assumed to be available. In convolution, the first element in a HW cipher needs to be added with some of its neighboring elements and before the addition, each of these elements need to be multiplied with different filter weights. To do so, we can left rotate each of these neighboring elements to the same position as the first element. Rotating the ciphertext vector in such a way moves the position appropriately for all HW elements. For each rotated vector, the same weight needs to be multiplied for all HW elements, so \himuls~of the ciphertext is sufficient. Before multiplication, the filter weight needs to be scaled using the compiler-provided scaling factor (plainLogP). These ciphertext vectors can then be added for different input channels to produce a ciphertext for an output channel. The rotations of the input ciphertexts are invariant to the output batch and channel, so they can be code motioned out in the implementation to reduce the number of rotations (omitted in the pseudocode for the sake of exposition). 

\paragraph{HW-tiled Homomorphic Convolution 2-d with SAME padding}: 
If a similar 2-d convolution has to be implemented, but with same padding, then each HW ciphertext is expected to be padded by the compiler so that the amount to rotate (left or right) varies but the number of rotations remain the same. However, after one convolution, there are invalid elements where the padded zeros existed earlier because those are added to the neighboring elements as well. The runtime keeps track of this using an additional metadata on the CipherTensor. The next time a convolution (or some other operation) is called on the CipherTensor, if the CipherTensor contains invalid elements where zeros are expected to be present, then the implementation can mask out all invalid elements with one \himulv~operation (the plaintext vector contains 1 where valid elements exist and 0 otherwise). This not only increases the time for the convolution operation, but it also increases the modulus Q required because \hidivs~may need to be called after such a masking operation. For security reasons, a larger Q can increase the N that needs to be used during encryption, thereby increasing the cost of all homomorphic operations.

\paragraph{CHW-tiled Homormorphic Convolution 2-d}:
Let us now consider a CHW-tiled CipherTensor that represents a 4-d tensor. To perform a 2-d convolution on this, even with VALID padding, \himulv~is required because different weights need to be multiplied to different channels that are all in the same cipher. In some FHE libraries like HEAAN, the asymptotic complexity of \himulv~is much higher than that of \himuls, thereby increasing the cost of the convolution operation. Moreover, after multiplication, the multiple input channels in the ciphertext need to be summed up into a single output channel. 
Such a reduction can be done by rotating every channel to the position of the first one in the ciphertext and adding them up one by one. If $C$ is the number of channels in the cipher, this involves $C-1$ rotations. However, such a reduction can be done more efficiently by exploiting the fact that the stride between the input channels is the same. This requires at the most $2log(C)$ rotations with additions in-between rotations, similar to vectorized reduction on unencrypted data. 
To produce an output ciphertext with multiple channels $C$, the input ciphertext needs to be replicated $C-1$ times. Instead of $C-1$ rotations serially, this can also be done in $2log(C)$ rotations, by adding the ciphertexts in-between.

\paragraph{Homomorphic matmul}:
Different FHE operations have different latencies. Although \hirotl~and \himulv~ have similar algorithmic complexity in HEAAN, the constants may vary and we observe that, \himulv~is more expensive than \hirotl. Due to this, it may be worth trading multiplications for rotations. This trade-off is most evident in homormorphic matmul operation. The number of \himulv~required reduce proportional to the number of replicas of the data we can add in the same cipher. Adding replicas increase the number of rotations but decrease the number of multiplications. This yields much more benefit because replicas can be added in log number of rotations instead of the linear number of multiplications.

%% file: sections/conv2d.tex
% !TEX root = ../main.tex
\begin{algorithm}[t]
\footnotesize
\caption{HW-Tiled Homomorphic Convolution 2-d of a 4-d tensor (with valid padding)}
\label{alg:conv2d}
\SetKwInOut{Input}{Input}
\SetKwInOut{Output}{Output}
\Input{\textsf{CipherTensor} input}
\Input{\textsf{PlainTensor} filter}
\Output{\textsf{CipherTensor} output}
\ForEach{[b, oc] in output.getOuterDims()} {
    output.ciphers[b, oc] = zeroCipher \\
    \ForEach{ic in input.getChannelDims()} {
        hStride, wStride = input.getCipherStrides() \\
        \ForEach{[fh, fw] in filter.getHeightWidth()} {
            weight = filter[fh, fw, ic, oc] \\
            weightFP = FixedPrecision(weight, plainLogP) \\
            rotate = fh * hStride + fw * wStride \\
            temp = input.ciphers[b, ic] \\
            FHE.\hirotlassign(temp, rotate) \\
            FHE.\himulsassign(temp, weightFP) \\
            FHE.\hiaddassign(output.ciphers[b, oc], temp)
        }
    }
    FHE.\hidivsassign(output.ciphers[b, oc], plainLogP)
}
\end{algorithm}

%% file: sections/compiler.tex
% !TEX root = ../main.tex    
\section{Compiler For Transforming Homomorphic Tensor Operations}
\label{sec:hec-compiler}

\begin{figure}
\centering
\includegraphics[width=0.35\textwidth]{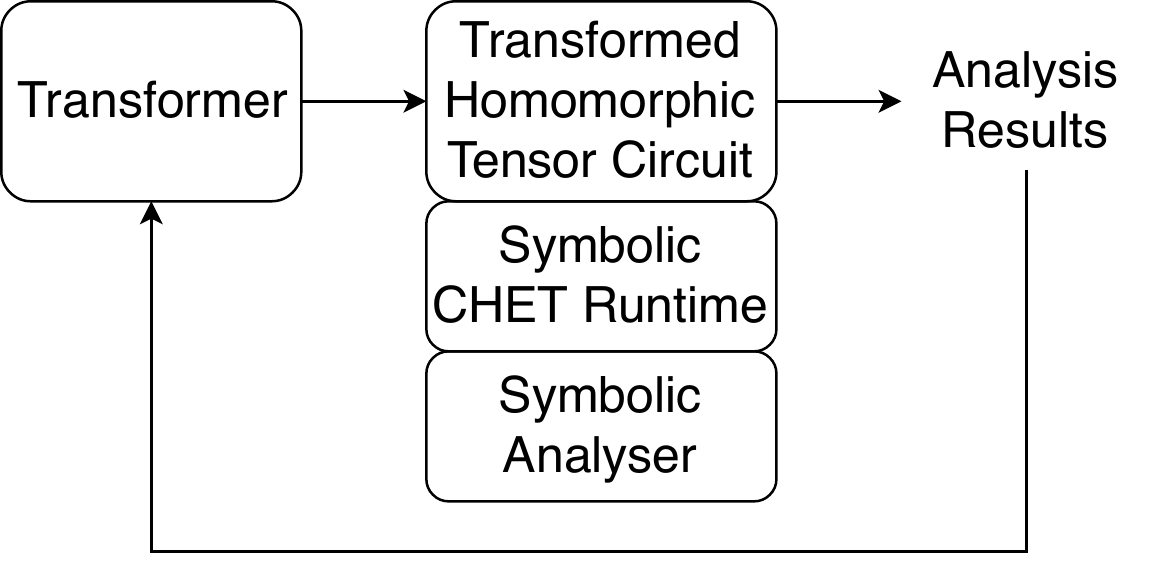}
\caption{Overview of an analysis and transformation pass in the \Hec compiler.}
\label{fig:compiler-pass}
\end{figure}

In this section, we describe the CHET compiler, which is a critical component of the end-to-end software stack to homomorphically evaluate tensor programs. The compiler is responsible for generating an optimized homomorphic tensor circuit that not only produces accurate results but also guarantees security of the data being computed. We describe the analysis and transformation framework used by the CHET compiler and then use this framework to describe a few transformations that are required for accuracy and security, and a few optimizations that improve performance significantly. We present the analysis and transformations specifically for HEAAN but it can extended for other FHE libraries trivially. 

\subsection{Analysis and transformation framework}
In contrast to most traditional optimizing compilers, the input tensor circuit has two key properties that the CHET compiler can exploit: the data flow graph is a Directed Acyclic Graph (DAG) and the tensor dimensions are known at compile-time from the schema provided by the user (similar to High Performance Fortran compilers). The CHET compiler must also analyze multiple data flow graphs corresponding to different implementations. Fortunately, the design space is not that large and we can explore it exhaustively, one at a time.

The CHET compiler only needs to determine the policy to execute homomorphic computation while the CHET runtime handles the mechanisms of efficient execution. This separation of concerns simplifies the code generation tasks of the compiler but it could complicate the analysis required. The compiler generates high-level homomorphic tensor operations but needs to analysis the low-level HISA instructions executed. To resolve this, we exploit the CHET runtime directly to perform the analysis.

Figure~\ref{fig:compiler-pass} shows the flow of a single analysis and transformation pass in the compiler. The transformer is a simple tool that specifies the parameters to use in the homomorphic tensor circuit (the input tensor circuit is directly mapped to an equivalent homomorphic tensor circuit with parameters left unspecified). This transformed homomorphic tensor circuit can be symbolically executed using the CHET runtime. This would execute the same HISA instructions that would be executed at runtime because the tensor dimensions are known. Instead of executing these HISA instructions using the FHE scheme, the HISA instructions invoke the symoblic analyser. This tracks the data that flows through the circuit. Repeated iteration to a fix-point is not needed since the circuit is a DAG. The DAG is unrolled on-the-fly to dynamically track the data flow, without having to construct an explicit DAG. Different analysers can track different information flows. The analyser returns its results as the output of the symoblic execution. The transformer then uses these results to instantiate a new, transformed specification of the homomorphic tensor circuit.

% necessary transformations
% optimizations

\subsection{Parameter Selection}
To choose encryption parameters such that the output is accurate, the compiler needs to analyse the depth of the circuit. In HEAAN, \hidivs~consumes modulus Q from the ciphertext it is called on. The output should have sufficient modulus Q left to capture accurate results. We implement a symoblic analyser with a dummy ciphertext datatype that increments the modulus Q of the input ciphertext and copies into the output ciphertext whenever \hidivs~is called. All other HISA instructions only copy the modolus from intput to output. The modulus Q in the dummy output ciphertext is the depth of the circuit. The input ciphertext should be encrypted with modulus Q that is at least the sum of the depth of the circuit and the desired output precision. This ensures accurate results. For the input modolus Q, a large enough N must be chosen to guaratee security of the data. This is a deterministic map from Q to N. In some cases, Q might require prohibitively large N, in which case the compiler needs to introduce bootstrapping in-between. 
% This is an NP-complete problem~\cite{bootstrapping-npcomplete}. 
We do not explore this in this paper.

\subsection{Padding Selection}
This analysis does not need to track the data flow in the HISA instructions. It requires analysing the metadata in the CipherTensor that flows through the homomorphic tensor circuit. This is quite straight-forward. If tensor operations like convolution need padding, then the previous operations until the input should maintain that padding. Some tensor operations may change strides, in which case the padding required scales by that factor. The input padding (in each ciphertext dimension) selected is the maximum padding required for any homomorphic tensor operation in the circuit.

\subsection{Optimization: Rotation Keys Selection}
By default, HEAAN inserts public evaluation keys for power-of-2 left and right rotations, and all rotations are performed using a combination of power-of-2 rotations. The ciphertext size is $N/2$, so HEAAN stores $2log(N)-2$ rotation keys by default. The rotation keys consume significant memory, so this is trade-off between space and performance. By storing only these keys, any rotation can be performed. However, this is too conservative. In a given homomorphic tensor circuit, the distinct slots to rotate would not be in the order of $N$. We use the analyser to track the distinct slots to rotate used in the homomorphic tensor circuit. All rotate HISA instructions store the constant slots to rotate in the analyser (right rotations are converted to left rotations) and other HISA instructions are ignored. The analyser returns the distinct slots to rotate that were used. The compiler generates the encryptor such that it would generate evaluation keys for these rotations using the private key on the client (Figure~\ref{fig:overview-runtime}). We do not need power-of-2 rotation keys and we observe that the rotation keys chosen by the compiler are a constant factor of $log(N)$.
% Possible to do completely because no control flow. With control flow (dependent
% on some plain text inputs) could still do static analysis to detect e.g. that
% rotations of a multiple of a stride are performed and insert power-of-two*stride
% rotation keys.

% \begin{figure}
% \centering
% \footnotesize
% \begin{tabularx}{0.72\columnwidth}{ll}
% \toprule
% Operation & Cost \\
% \midrule
% \hiadd*, \hisub* & $O(n)$ add/sub \\ 
% \himuls*, \hidivs* & $O(n)$ mult \\
% \himulv*, \himul, \himulassign & $O(n\log(n)) $ mult \\
% \hirelin, \hirotl*, \hirotr* & $O(n \log (n))$ mult \\
% \bottomrule
% \end{tabularx}
% \caption{HEAAN asymptotic costs. Unit: operations in integers mod $Q$}
% \label{fig:heaan-cost}
% \end{figure}

\subsection{Optimization: Data Layout Selection}
Different homomorphic operations have different costs, even within the same FHE scheme. 
% Figure~\ref{fig:heaan-cost} lists the asymptotic complexity of different homomorphic operations in HEAAN. 
The cost of homomorphic operations could also vary a lot based on small variations of the scheme. For example, by using a full-RNS variant of the HEAAN scheme and storing the ciphertexts and plaintexts in the number-theoretic-transform (NTT) domain, we can decrease the complexity of \himulv* from $O(n\log(n))$ to $O(n)$. The compiler can encode the cost of each operation either from asymptotic complexity or from microbenchmarking each operation. These costs can then be used to decide which implementation or data layout to choose for a given homomorphic tensor circuit. The analyser in this case counts the number of occurrences of each operation and then uses the asymptotic complexity to determine to the total cost of the circuit. The transformer creates a homomorphic tensor circuit corresponding to a data layout and gets its cost using the analyser. It repeats this for different possible data layout options and then chooses the one with the lowest cost.

%% file: sections/evaluation.tex
% !TEX root = ../main.tex
\section{Evaluation}
\label{sec:eval}

\newcommand{\mnistall}{LeNet-5-like\xspace}
\newcommand{\mnistsmall}{LeNet-5-small\xspace}
\newcommand{\mnistmed}{LeNet-5-medium\xspace}
\newcommand{\mnistlarge}{LeNet-5-large\xspace}
\newcommand{\mnistlargehf}{LeNet-5-large-hf\xspace}
\newcommand{\mnetname}{Industrial\xspace}
\newcommand{\squeezenet}{SqueezeNet-CIFAR\xspace}

Our evaluation targets a set of neural network architectures for image
classification tasks described in Figure~\ref{fig-all-nets}.

\begin{figure}
\footnotesize
\begin{tabularx}{\columnwidth}{@{}X|rrr|r@{}}
\toprule
 & \multicolumn{3}{c|}{No. of layers} & \\
Network & Conv & FC & Act &  \# FP operations\\
\midrule
\mnistsmall & 2 & 2 & 4 & 159960 \\
\mnistmed & 2 & 2 & 4 & 5791168 \\
\mnistlarge & 2 & 2 & 4 & 21385674 \\
\mnetname & 5 & 2 & 6 & - \\
\squeezenet & 10 & 0 & 9 & 37759754 \\
\bottomrule
\end{tabularx}
\caption{DNNs used in our evaluation.}
\label{fig-all-nets}
\vspace{-5pt}
\end{figure}

\begin{description}
\item[\mnistall] is a series of networks for the MNIST~\cite{mnist}
dataset. We use three versions with different number of neurons:
\mnistsmall, \mnistmed, and \mnistlarge. The largest one matches the
one used in the TensorFlow's tutorials~\cite{lenet5like}. These networks
have two convolutional layers, each followed by ReLU activation and max
pooling, and two fully connected layers with a ReLU in between. 
% The full network structure for \mnistlarge can be seen in Figure~\ref{fig-lenet-5-like}.
\item[\squeezenet] is a network for the CIFAR-10 dataset~\cite{cifar10}
that follows the SqueezeNet~\cite{SqueezeNet} architecture. This version has
4 Fire-modules~\cite{squeezenet-cifar10} for a total of 10 convolutional
layers. 
% Please see Figure~\ref{fig-squeezenet-cifar10} for the full network
% structure. Layers 11 and 20 include the simple residual connections described
% in~\cite{SqueezeNet}.
\item[\mnetname] is an pretrained HE-compatible network from an industry partner
for a privacy-sensitive image classification task. We are unable to
reveal the details of the network other than the fact that it has 5
convolutional layers and 2 fully connected layers. 
\end{description}

All networks other than \mnetname use ReLUs and max-pooling,
which are not compatible with homomorphic evaluation. For these
networks, we modified the activation functions to a second-degree
polynomial~\cite{cryptonets,Chabanne2017}. The key difference with
prior work is that our activation functions are $f(x) = ax^2 + bx$
with learnable parameters $a$ and $b$. During the training phase, the
DNN adjusts these parameters automatically to appropriately
approximate the ReLU function. To avoid exploding the gradients during
training (which usually happens during the initial parts of training),
we initialized $a$ to zero and clipped the gradients when large. We
also replaced max-pooling with average-pooling. 

% For \mnistlarge we additionally quantized the weights to 8-bit fixed
% point precision and changed the padding. We obtained a training
% accuracy of 0.993 which matches the accuracy of the base network. 

%We trained ``LeNet-5-like'' to an accuracy of 0.993. For the HE-compatible
%version, we replaced ReLUs with $f(x)=ax^2+bx$ and max-pooling with
%average-pooling, which resulted in the same accuracy of 0.993. Further
%introducing quantization to a 8-bit fixed point precision and changing to valid
%padding did not degrade accuracy from 0.993.

We trained \mnistlarge to an accuracy of 0.993, which matches that of the network in 
the TensorFlow's tutorial~\cite{lenet5like}. 
For \squeezenet, we additionally introduce L2-regularization with a scale of
$1\times{}10^{-5}$ to improve performance. Our resulting accuracy is
0.815 which is close to the accuracy of 0.84 of the non-HE compatible
model. To the best of our knowledge \squeezenet is the deepest
neural network that has been homomorphically evaluated.

%As noted in~\cite{cryptonets}, using square activation functions can introduce
%exploding gradients during, especially the initial parts of, training. We manage
%this problem by initializing the $a$ weight in our activation functions to zero
%and clipping the norm of gradients during training.

All experiments were run on a dual socket Intel Xeon E5-2667v3@3.2GHz with 224
GiB of memory. Hyperthreading was off for a total of 16 hardware threads. All
runtimes are reported as averages over 20 different images.
We present the average latency of image inference with a batch size of 1.

\begin{figure}[t]
\footnotesize
\begin{tabularx}{\columnwidth}{@{}Xrr@{}}
\toprule
Model & CHET & Hand-written \\
\midrule
\mnistsmall & 8 & 14 \\
\mnistmed & 51 & 140 \\
\mnistlarge & 265 & - \\
%\mnistlargehf & 211  & - \\
\mnetname & 312 & 2413 \\
SqueezeNet-CIFAR10 & 1342 & - \\
\bottomrule
\end{tabularx}
\caption{Average latency (in seconds) of CHET and hand-written versions.}
\label{fig:hand}
\end{figure}

\begin{figure}[t]
\footnotesize
\begin{tabularx}{\columnwidth}{@{}Xrrrrr@{}}
\toprule
Model & $\operatorname{log}(N)$ & $\operatorname{log}(Q)$ &
$\operatorname{log}(P_\mathit{c})$ &
$\operatorname{log}(P_\mathit{p})$ \\ %& $\operatorname{log}(P_\mathit{m})$ \\
\midrule
\mnistsmall & 14 & 240 & 30 & 16 \\ %& 15 \\ %& 8192 \\
\mnistmed & 14 & 240 & 30 & 16 \\ %& 15 \\ %& 8192 \\
\mnistlarge & 15 & 400 & 40 & 20 \\ %& 20 \\ %& 16384 \\
%\mnistlargehf & 15 & 380 & 40 & 20 \\ %& 20 \\ %& 16384 \\
\mnetname & 16 & 705 & 35 & 25 \\ %& 20 \\ %& 32768 \\
SqueezeNet-CIFAR10 & 16 & 940 & 30 & 20 \\ %& 20 \\ %& 32768 \\
\bottomrule
\end{tabularx}
\caption{Encryption parameters selected by CHET and the user-provided precisions for each model.}
\label{fig:params}
\vspace{-5pt}
\end{figure}

\begin{figure}[t]
\centering
\footnotesize
\begin{tabularx}{\columnwidth}{@{}Xrrrr@{}}
\toprule
\multirow{2}{*}{Model} & \multirow{2}{*}{HW} & \multirow{2}{*}{CHW} & HW-conv & CHW-fc \\
& & & CHW-rest & HW-before \\
\midrule
\mnistsmall & 8 & 12 & 8 & \textbf{8} \\
\mnistmed & 82 & 91 & 52 & \textbf{51} \\
\mnistlarge & 325 & 423 & 270 & \textbf{265} \\
%\mnistlargehf & 218 & 281 & 216 & \textbf{211} \\
\mnetname & 330 & \textbf{312} & 379 & 381 \\
\squeezenet & \textbf{1342} & 1620 & 1550 & 1342 \\
\bottomrule
\end{tabularx}
\caption{Average latency (in seconds) with different layouts.}
\label{fig:tilings}
\end{figure}

\begin{figure}[t]
\centering
\footnotesize
\begin{tabularx}{\columnwidth}{@{}Xrr@{}}
\toprule
Model & Unoptimized & Optimized \\
\midrule
\mnistsmall & 14 & 8 \\
\mnistmed & 73 & 51 \\
\mnistlarge & 426 & 265 \\
%\mnistlargehf & 300 & 211 \\
\mnetname & 645 & 312 \\
\squeezenet & 2648 & 1342 \\
\bottomrule
\end{tabularx}
\caption{Average latency (in seconds) with and without rotation keys optimization.\label{fig-rotopt}}
\vspace{-5pt}
\end{figure}

% \begin{figure}[t]
% \centering
% \footnotesize
% \begin{tabularx}{0.4\textwidth}{lrrr}
% \toprule
% Model & MoreMultLessRot & NoLogRot & Best \\
% \midrule
% \mnistsmall & & 9 & 8 \\
% \mnistmed & & 60 & 51 \\
% \mnistlarge & & 310 & 265 \\
% \mnistlargehf & & 245 & 211 \\
% \mnetname & & 435 & 312 \\
% % \squeezenet & & & \\
% \bottomrule
% \end{tabularx}
% \caption{Average latency (in seconds) with different implementations of the runtime (with all compiler optimizations).}
% \end{figure}

\paragraph{Comparison with hand-written:}
Figure~\ref{fig:hand} compares hand-written implementations and CHET with all optimizations. CHET clearly outperforms hand-written implementations. The hand-written implementations lack some of the optimizations in the CHET compiler and runtime. Moreover, it is difficult to scale these hand-written implementations to large networks like \mnistlarge and SqueezeNet-CIFAR10, so we do not have hand-written implementations for these to compare against.

\paragraph{Parameter Selection:}
In Figure~\ref{fig:params}, the last columns show precision (the number of decimal digits) required for the image or ciphertext ($P_c$) and the weights or the plaintext ($P_p$), that are provided by the user. The precision provided is used by the CHET compiler to select the encryption parameters $N$ and $Q$. The values of these parameters grow with the depth of the circuit, as shown in the figure.
With these parameters, CHET generated homomorphic tensor circuits achieve the same accuracy as the unencrypted circuits. In addition, the difference between the output values of these circuits is within the desired precision of the output.

\paragraph{Data Layout Selection:} 
We evaluate four different data tiling layouts choices for the three largest networks: 
(i) HW: each ciphertext has all height and width elements of a single channel,
(ii) CHW: each ciphertext has multiple channels (all height and width elements of each),
(iii) HW-conv and CHW-rest: same has CHW, but move to HW before each convolution and back to CHW after each convolution, and
(iv) CHW-fc and HW-before: same as HW, but switch to CHW during the first fully connected layer and CHW thereafter.
Figure~\ref{fig:tilings} presents the average latency of each network for each
layout. We can see that for each network, a different layout provides the
lowest latency. It is very difficult for the user to determine which is the best data layout and more importantly, it is difficult to implement each network manually using a different data layout. This highlights how the compiler should
search the space of possible layouts and kernel implementations for each program
separately, while entrusting the runtime to implement it efficiently. In this case, the compiler chooses the best performing data layout for each network based on the cost model of HEAAN.

\paragraph{Rotation Keys Selection:}
We evaluate the efficacy of our rotation keys optimization on the three largest networks.
Figure~\ref{fig-rotopt} presents the average latencies with the optimization on
or off. The optimization provides significantly improved performance for all
networks and should be always used.\footnote{Barring very memory constrained
environments.} Having this optimization implemented as an automatic compiler
pass removes the burden of adding proper rotation keys in each program
separately.

%% file: sections/related.tex
\section{Related Work}
FHE is currently an active area of research. See Acar et
al~\cite{fhe-survey} for more details. Many have observed the need and
suitability of FHE for machine learning tasks. We survey the most
related work below. 

Cryptonets~\cite{cryptonets} was the first tool to demonstrate a
fully-homomorphic inference of a DNN by replacing the (FHE
incompatible) RELU activations with a quadratic function. They
demonstrated an accuracy of 98.95\% for MNIST with a network with two
hidden layers, with a latency of 250 seconds per image. 
%Their encoding
%allowed a parallel inference of 4096 images at the same time,
%giving them a throughput of 59000 predictions per hour. However, they
%report that training the network with quadratic activations are
%unstable because of its unbounded gradient. 
Chabanne et al.~\cite{Chabanne2017} improve upon Cryptonets by using
batch normalization~\cite{IoffeS15} to bound the inputs to the
activations into a small interval where the quadratic approximation is
valid. They report improvement in accuracy
for MNIST to 99.30\% with a network with 6 activation layers. Our
method of retraining the quadratic activation function is inspired by
these works. Our emphasis in this paper is primarily on automating the manual
and error-prone hand tuning required to ensure that the networks are
secure, correct, and efficient. 

Bourse et al.~\cite{tfhe-mnist} use the TFHE library~\cite{TFHE} for
DNN inference. TFHE operates on bits and is thus slow for multi-bit
integer arithmetic. To overcome this difficulty, Bourse et al. instead
use a discrete binary neural network (where activation functions output either -1
or 1) with a hidden layer using the sign function as the activation.  
Our goal in this paper is to build a framework for compiling
larger general-purpose DNNs. 

Similarly, Cingulata~\cite{armadillo, cingulata} is a compiler for
converting C++ programs into a Boolean circuit, which is then
evaluated using a backend FHE library. Despite various
optimizations~\cite{cingulata-opts}, this approach is unlikely to
scale for large DNNs.

DeepSecure~\cite{RouhaniRK17} and SecureML~\cite{secureml} use
secure multi-party computation techniques to respectively perform DNN inference and
training. Such techniques employ communication between multiple
entities that are assumed to not collude and thus do not have the
simple trust model of FHE. On the other hand, such approaches are more
flexible in the operations they can perform and are computationally
cheaper. 

Finally, prior work~\cite{mrcrypt,jcrypt} have used partially
homomorphic encryption schemes (which support addition or
multiplication of encrypted data but not both) to determine the
encryption schemes to use for different data items so as to execute a
given program. While they are able to use computationally efficient
schemes, such techniques are not applicable for evaluating DNNs that
require both multiplication and addition to be done on the same
input. 

%\paragraph{Partial Homomorphic Encryption}

%https://ieeexplore.ieee.org/document/6514926/

%http://sirdeyre.free.fr/Papiers_etc/2015_Armadillo_A_compilation_chain_for_privacy_preserving_applications.pdf: C++ to Boolean gates

%autocrypt: enclaves https://dl.acm.org/citation.cfm?id=2541806.2516666

%CryptDB: https://dl.acm.org/citation.cfm?id=2043566

%DSL for PHE: https://eprint.iacr.org/2011/561.pdf

%http://www.cs.rpi.edu/~milanova/docs/HoTSoS18.pdf

%\end{verbatim}

%% file: sections/conclusion.tex
\section{Conclusion}

Good abstractions separate concerns to either side of that
abstraction. For example, \texttt{x86} lets hardware manufacturers
innovate independently of the software developers and compiler writers
that target that abstraction.  Likewise, \texttt{cuDNN} lets machine
learning experts target various hardware implementations and reap the
benefits of new GPU hardware without changing their software.  

This paper introduces such an abstraction for FHE applications that
cleanly abstracts and exposes features of FHE implementations. We
demonstrate a compiler and runtime that targets FHE tensor programs,
evaluate that compiler and runtime on real-world CNN models, and
demonstrate our compiler is able to significantly optimize the
performance of FHE tensor programs.  Because of these optimizations,
this paper demonstrates the deepest FHE based CNNs to date.

This paper demonstrates the fruitful application of systems and
compiler research for FHE applications. We posit there are many open
problems that such areas can help solve (i.e., compiler support that
turns expressions into FHE compatible ones, or more optimization
passes specific to encryption parameters).  We expect these opportunities
excites other researchers as much as it does the authors.